\documentclass{esannV2}
\usepackage[dvips]{graphicx}
\usepackage[latin1]{inputenc}
\usepackage{amssymb,amsmath,array}
\usepackage{subfig}

\begin{document}
\title{From User-independent to Personal Human Activity Recognition Models Exploiting the Sensors of a Smartphone}

\author{Pekka Siirtola, Heli Koskim\"{a}ki and Juha R\"{o}ning
%
%
\vspace{.3cm}\\
Biomimetics and Intelligent Systems Group,\\ P.O. BOX 4500, FI-90014, University of Oulu, Oulu, Finland\\
pekka.siirtola, heli.koskimaki, juha.roning@ee.oulu.fi
}

\maketitle


%
%
%
\hyphenation{manu-scripts manu-script ext-re-mums user-in-de-pen-dent user-de-pen-dent in-de-pen-dent mo-del in-de-pen-dent}

\begin{abstract}

In this study, a novel method to obtain user-dependent human activity recognition models unobtrusively by exploiting the sensors of a smartphone is presented. The recognition consists of two models: sensor fusion-based user-independent model for data labeling and single sensor-based user-dependent model for final recognition.
The functioning of the presented method is tested with human activity data set, including data from accelerometer and magnetometer, and with two classifiers. Comparison of the detection accuracies of the proposed method to traditional user-independent model shows that the presented method has potential, in nine cases out of ten it is better than the traditional method, but more experiments using different sensor combinations should be made to show the full potential of the method.

\end{abstract}

\section{Problem statement and related work}
\label{problem}

Smartphones have become increasingly common and people carry them everywhere they go, this means that nowadays almost every one has a device capable to be used to recognize activities. This of course enables a lot of possilities to the field of activity recognition.
As known, smartphones include a wide range of sensors, such as magnetometer, gyroscope, GPS, proximity sensor, ambient light sensor, thermometer and barometer. Still, most activity recognition studies use only accelerometers in detection, for instance \cite{khan}, \cite{kose}, \cite{liang}, \cite{ARready}. 
However, by fusing the sensors of a mobile phone more accurate models could be build than using only one sensor \cite{Shoaib}. On the other hand, this improvement comes at the expense of energy efficiency. While smartphones have a limited battery capacity, the sensor fusion-based recognition is seldom a usable approach when the aim is to monitor everyday life 24/7. This calls for innovative solutions to employ sensor fusion. For instance, in \cite{Wang} a smart and energy-efficient way to deploy the sensors of a mobile phone to recognize activities was presented. The method presented in the study uses the minimum number of sensors needed to detect user's activity reliably and when activity changes, more sensors are used to detect the new activity. By using this type of smart sensor selection, the battery life can be improved by 75\%. In addition, in \cite{Altini} the idea to use sensor fusion to build adaptive recognition models was presented. In the study, a method to personalize user-independent walking speed estimation model is presented. In the study, it is noted that user-independent walking speed estimation model based on accelerometer data is not accurate when walking in unconstrained conditions. Therefore, the study introduced an automatic calibration methodology  combining accelerometer and GPS data to find a person-specific offset to be used with user-independent estimation model. Offset was determined by comparing walking speed estimation at treadmill to speed measured by the GPS outdoors. By using this method, it was possible to reduce the walking speed estimation error by 8.8\%

The idea of the novel method presented in this study is to build user-dependent recognition models without need for a separate data collection session. Typically data collecting phase is compulsory, and therefore user-independent methods are often preferred instead of personal ones, though, it is shown that user-dependent recognition models are more accurate than user-independent \cite{weiss2012impact}.

\section{Experimental dataset}
\label{data}


The data used in this study is already used in \cite{ARready}. However, in this study only data collected from trousers' front pocket was used. Data was available from five subjects and these subjects performed five different activities: walking, running, cycling, driving a car, and idling, that is, sitting/standing. 

The data were collected using a Nokia N8 smartphone running Symbian\textasciicircum3 operating system. N8 includes several sensors, however, in this study only accelerometer and magnetometer were used. The used sampling rate was 40Hz. The total amount of the data was about fifteen hours.

%

\section{From user-independent to personal recognition models}
\label{aim}

\begin{figure}[t]
	\centering
		\includegraphics[width=8.7cm,keepaspectratio = true]{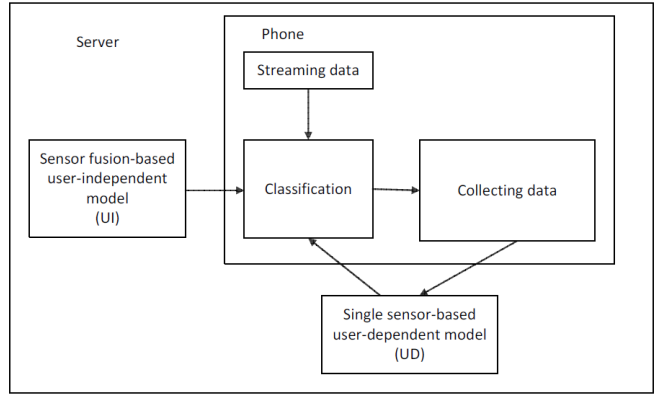}
\caption{The proposed method consists of two classification models: user-independent model which tries to maximize the recognition accuracy, and user-dependent model which minimizes the computational load.}
\label{ucrp}
\end{figure}

The proposed method is presented in Figure \ref{ucrp} and it consists of four phases:

\textit{1. Train sensor fusion-based user-independent model.} In the first phase, sensor fusion-based recognition model is used to recognize activities from a streaming data. To maximize the recognition rate of this model, it is trained using a large number of features. These can include for instance features extracted from time domain, as well as, frequency domain. Moreover, these features can exploit more than one type of sensors of a smartphone.

\textit{2. Collect and label personal data when user is using the recognition application based on user-independent model.} When streaming data is classified using sensor fusion-based user-independent model, it can be assumed that recognition is reliable. Therefore, by combining these recognition results, and using them as labels, and the data related to them, it is possible to collect personal training data set while sensor fusion-based model is used to recognize activities.

\textit{3. Train the single sensor-based user-dependent model.} When personal data from each of the recognized activities is available, user-dependent recognition model can be trained based on it using the classification results as labels. In order to make this personal recognition model light, it needs to be based on a low number of features extracted from a data of one sensor and from one domain. 
In practice, it is wise to built accelerometer-based user-dependent model, as accelerometers are the most energy efficient sensors, and most accurate as well. 

\textit{4. Recognize activities using the user-dependent model.} Streaming data can now be classified using a light, single sensor-based user-dependent model.

\section{Methods}
\label{method}


In this study, the effect of gravitation was eliminated from the sensor readings by combining all three acceleration channels as one using sum of squares. The same was done to magnetometer signals as well. Moreover, calibration differences between devices were eliminated by using the method presented in \cite{ARready}.

%

In the feature extraction, sliding window method was used with window length of 1 second, which is equal to 40 samples as the used sampling rate was 40Hz, and slide of 0.25 seconds.
To train the single sensor-based user-dependent model, 19 features were extracted from magnitude acceleration signal.
These features were standard deviation, minimum, maximum. In addition, instead of extracting percentiles, the remainder between percentiles (10, 25, 75, and 90) and median were calculated. Moreover, the sum of values above or below percentile (10, 25, 75, and 90), square sum of values above of below percentile (10, 25, 75, and 90), and number of crossings above or below percentile (10, 25, 75, and 90) were extracted and used as features.  The user-independent model uses these same features as well, but they are extracted from both magnitude acceleration signal and magnitude magnetometer signal. In addition, from these signals frequency domain features were extracted. These features included sums of smaller sequences of Fourier-transformed signals. This way, user-independent model was trained using altogether 56 features. 

The most descriptive features for each model were selected using a SFS method (sequential forward selection, \cite{devijver82}). Moreover, to reduce the number of misclassified windows, the final classification was done based on the majority voting of the classification results of three adjacent windows. Therefore, when an activity changes, a new activity can be detected when two adjacent windows are classified as a new activity.


Experiments were done using two classifiers to be able to compare how the proposed method works with different classifiers. These classifiers were  linear discriminant analysis (LDA), and quadratic discriminant analysis (QDA) as in our previous studies we have noticed they are not only accurate but also computationally light, and therefore, sufficient to be implemented to smartphones and used 24/7.



\section{Experiments}
\label{experiments}

\begin{figure}[t]
	\centering
		\includegraphics[width=8.5cm,keepaspectratio = true]{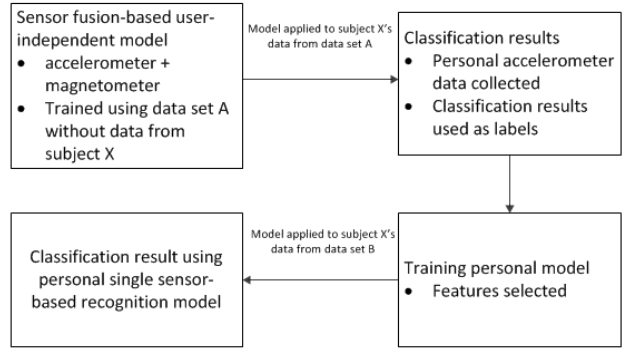}
\caption{To test the accuracy of the proposed method, two data sets from each test person is needed. This figure shows how these data sets are used.}
\label{exp1}
\end{figure}

\begin{figure}[t]
	\centering
		\includegraphics[width=8.5cm,keepaspectratio = true]{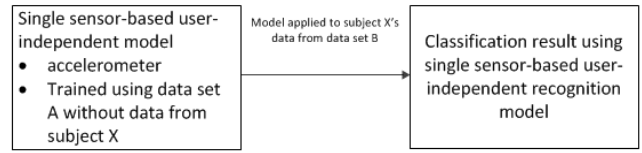}
\caption{The proposed method was compared to traditional user-independent recognition.}
\label{exp2}
\end{figure}

\begin{table*}[t]
\begin{footnotesize}
\caption{Activity recognition rates of user-dependent models for each subject.}
\begin{center}
\begin{tabular}{|p{1.6cm}|p{1.2cm}|p{1.2cm}|p{1.2cm}|p{1.2cm}|p{1.2cm}|p{0.9cm}|}

\hline

 \textbf{Accuracy} &  \textbf{Subj 1} &  \textbf{Subj 2} &  \textbf{Subj 3} &  \textbf{Subj 4} &  \textbf{Subj 5}&  \textbf{Avg}\tabularnewline
\hline
\textbf{LDA} &  & & & & &\tabularnewline
proposed &  77.7\%& 80.8\%& \textbf{89.5\%}& 76.9\%& \textbf{86.0\%}& 82.2\%\tabularnewline
traditional &  76.5\%& 74.5\%& 87.7\%& 73.0\%& 83.3\%& 79.0\%\tabularnewline
improvement &  1.6\%& 8.5 \%& 2.1 \%& 5.3 \%& 3.2\%&4.1\%\tabularnewline
\textbf{QDA} &  & & & & &\tabularnewline
proposed &  79.6\%& \textbf{83.6}\%& 87.8\%& \textbf{78.3\%}& 83.0\%& \textbf{82.5\%}\tabularnewline
traditional &  \textbf{79.8\%}& 82.5\%& 86.8\%& 75.5\%& 73.6\%& 79.6\%\tabularnewline
improvement &  -0.3\%& 1.3 \%& 1.2\%& 3.7 \%& 12.8\%&3.6\%\tabularnewline
\hline
\end{tabular}

\label{data2}
\end{center}
\end{footnotesize}
\end{table*}

%
%

The proposed method was tested using the experiment protocol presented in Figure \ref{exp1}. For this purpose, the data from each subject was divided into half, one half was used for training (referred as data set A in Figure \ref{exp1}) and the other half for testing (referred as data set B in Figure \ref{exp1}). The recognition rates are then calculated using leave-one-out method. One subject's data in turn was used as validation data, and as explained in the Figures \ref{exp1} and \ref{exp2}, training and test data were used to select features and train the recognition model.
 

The results are presented in Table \ref{data2}. The shown accuracies are detection rates of validation data, which was not used in feature selection or model training process making it totally unknown to the recognition model. In addition, the accuracies are obtained by calculating average of class-wise detection rates.

\section{Discussion}
\label{discussion}

The presented method improves classification accuracy in comparison to traditional single sensor-based user-independent model (Table \ref{data2}). In nine cases out of ten the proposed method improves the recognition accuracy. However, on average the improvement is only modest, from 3 to 4 \%.

\begin{table*}[t]
\begin{footnotesize}
\caption{Detection rates of the sensor fusion-based user-independent classifier.}
\begin{center}
\begin{tabular}{|p{1.3cm}|p{1.2cm}|p{1.2cm}|p{1.2cm}|p{1.2cm}|p{1.2cm}|p{0.8cm}|}

\hline
 \textbf{Accuracy} &  \textbf{Subj 1} &  \textbf{Subj 2} &  \textbf{Subj 3} &  \textbf{Subj 4} &  \textbf{Subj 5}&  \textbf{Avg}\tabularnewline
\hline
\textbf{LDA}& 85.4\%& 79.9\%& 93.0\%& 84.9\%& 85.3\%& 85.6\%\tabularnewline
\textbf{QDA}&  86.2\%& 82.2\%& 90.2\%& 87.1\%& 87.4\%& 86.6\%\tabularnewline
\hline
\end{tabular}

\label{model1}
\end{center}
\end{footnotesize}
\end{table*}


According to our literature study, the most similar previous work compared to this study is \cite{Altini}, where a method to personalize a model to estimate walking speed was presented. This reduced the walking speed estimation error by almost ten percent. When this improvement is compared to improvements gained in this study (Table \ref{data2}), it can be noted that they are lower. 
However, while the method presented in \cite{Altini} is quite similar to one presented in this study, here it is applied in different way and to different problem. This explains the differences in improvement percentages. Moreover, it is shown in \cite{weiss2012impact} that personal models can be over 20\%-units more accurate than user-independent models. In order to obtain such improvements, the sensor fusion-based model should be trained using more sensors. This would ensure that the data set used to train the personal model has less incorrect labels, which would lead to more accurate user-dependent models. The labels used in this study were not not as accurate as they were suppose to be which can be seen from Table \ref{model1} showing the subject-wise recognition rates of the sensor fusion-based user-independent model.

\section{Conclusions}
\label{conclusion}

In this study, a method to obtain light weight user-dependent human activity recognition models unobtrusively by exploiting the sensors of a smartphone was presented. The preliminary results are promising, in nine cases out of ten the proposed method improves the recognition accuracy. However, the method still requires work, e.g. it needs to be tested with more data sets and sensors. In addition, online implementation is a part of the future work.


\begin{thebibliography}{1}

\bibitem{khan}
A.~M. Khan, M.~H. Siddiqi, and S-W Lee.
\newblock Exploratory data analysis of acceleration signals to select
  light-weight and accurate features for real-time activity recognition on
  smartphones.
\newblock {\em Sensors}, 13(10):13099--13122, 2013.

\bibitem{kose}
M.~Kose, O.D. Incel, and C.~Ersoy.
\newblock Online human activity recognition on smart phones.
\newblock In {\em Workshop on mobile sensing: from smartphones and wearables to
  big data (colocated with IPSN)}, pages 11--15, 2012.

\bibitem{liang}
Y.~Liang, X.~Zhou, Z.~Yu, and B.~Guo.
\newblock Energy-efficient motion related activity recognition on mobile
  devices for pervasive healthcare.
\newblock {\em Mobile Networks and Applications}, pages 1--15, 2013.

\bibitem{ARready}
P~Siirtola and J.~R\"{o}ning.
\newblock Ready-to-use activity recognition for smartphones.
\newblock In {\em IEEE Symposium on Computational Intelligence and Data Mining
  (CIDM 2013)}, April 16--19 2013.

\bibitem{Shoaib}
M.~Shoaib, S.~Bosch, O.~D. Incel, H.~Scholten, and P.~Havinga.
\newblock Fusion of smartphone motion sensors for physical activity
  recognition.
\newblock {\em Sensors}, 14(6):10146--10176, 2014.

\bibitem{Wang}
S.~Wang, C.~Chen, and J.~Ma.
\newblock Accelerometer based transportation mode recognition on mobile phones.
\newblock In {\em Wearable Computing Systems (APWCS), 2010 Asia-Pacific
  Conference on}, pages 44 --46, 2010.

\bibitem{Altini}
M.~Altini, R.~Vullers, C.~Van~Hoof, M.~van Dort, and O.~Amft.
\newblock Self-calibration of walking speed estimations using smartphone
  sensors.
\newblock In {\em Pervasive Computing and Communications Workshops (PERCOM
  Workshops), 2014 IEEE International Conference on}, pages 10--18, March 2014.

\bibitem{weiss2012impact}
G.~Weiss and J.~Lockhart.
\newblock The impact of personalization on smartphone-based activity
  recognition.
\newblock In {\em AAAI Workshop on Activity Context Representation: Techniques
  and Languages}, 2012.

\bibitem{devijver82}
P.~A. Devijver and J.~Kittler.
\newblock {\em {Pattern recognition: A statistical approach}}.
\newblock Prentice Hall, 1982.

\end{thebibliography}
\end{document}